\documentclass{article}

 \usepackage[preprint]{neurips_2026}

\usepackage[utf8]{inputenc} 
\usepackage[T1]{fontenc}    
\usepackage{hyperref}       
\usepackage{url}            
\usepackage{booktabs}       
\usepackage{amsfonts}       
\usepackage{nicefrac}       
\usepackage{microtype}      
\usepackage{xcolor}         
\usepackage{graphicx}       
\usepackage{soul}
\usepackage{siunitx}
\usepackage{booktabs}
\usepackage{colortbl}
\usepackage{amsmath}
\usepackage{mathtools}
\usepackage{subcaption}
\usepackage{fontawesome5}

\usepackage{xcolor}
\definecolor{taskcolor}{HTML}{3B82F6}   
\newcommand{\task}[1]{\textcolor{taskcolor}{\texttt{#1}}}

\usepackage[table]{xcolor} 
\definecolor{lightblue}{RGB}{230, 240, 255} 

\usepackage{multirow}
\usepackage[table]{xcolor}
\definecolor{camerared}{RGB}{255, 220, 220}
\definecolor{procblue}{RGB}{220, 235, 255}
\definecolor{gengreen}{RGB}{220, 245, 220}
\definecolor{uiorange}{RGB}{255, 230, 205}
\definecolor{cameraredtext}{RGB}{180, 50, 50}
\definecolor{procbluetext}{RGB}{30, 80, 180}
\definecolor{gengreentext}{RGB}{30, 120, 60}
\definecolor{uiorangetext}{RGB}{180, 80, 20}

\providecolor{rgbvidcol}{HTML}{C0392B}
\providecolor{modvidcol}{HTML}{808080}
\providecommand{\rgbvid}{{\color{rgbvidcol}\faVideo}}
\providecommand{\modvid}[1]{#1\,{\color{modvidcol}\faVideo}}
\providecommand{\rgbimg}[1]{#1\,{\color{rgbvidcol}\faImage}}
\providecommand{\modimg}[1]{#1\,{\color{modvidcol}\faImage}}

\usepackage{siunitx}

\usepackage{xcolor}
\usepackage{amssymb}
\usepackage{xspace}
\usepackage{array}
\definecolor{tilegray}{HTML}{E8E8E8}
\definecolor{tiletarget}{HTML}{FFE8B0}

\definecolor{demoinput}{RGB}{255,100,78}
\definecolor{demotarget}{RGB}{254,174,0}
\definecolor{queryinput}{RGB}{96,217,55}
\definecolor{querytarget}{RGB}{45,144,255}

\newcommand{\ca}{\ensuremath{{\color{demoinput}A}}\xspace}
\newcommand{\cb}{\ensuremath{{\color{demotarget}B}}\xspace}
\newcommand{\cc}{\ensuremath{{\color{queryinput}C}}\xspace}
\newcommand{\cd}{\ensuremath{{\color{querytarget}D}}\xspace}

\newcommand{\model}{\textbf{\textsc{ViGeo}}}

\title{
Unifying Video Tasks via Spatiotemporal Analogy
}

\author{%
  Chia-Hsiang Kao$^{1}$\thanks{Corresponding author: ck696@cornell.edu} \quad
  Belinda Zeng$^{2}$ \quad
  Bharath Hariharan$^{1}$ \quad
  Menglin Jia$^{2}$ \\[0.5em]
  \small $^{1}$Cornell University \qquad $^{2}$Meta \\
}

\begin{document}

\maketitle
\begin{abstract}
Adapting video models to new tasks typically requires dedicated data curation and fine-tuning.
While visual analogy provides a training-free alternative by specifying tasks in-context, it remains restricted to the image domain.
To explore whether analogy-based methods can unify diverse video tasks and generalize to out-of-distribution scenarios, we introduce \model, a framework that extends visual in-context learning to the video domain via spatiotemporal canvas completion.
Evaluated on a diverse task taxonomy with a strict train-test split, \model~generalizes to unseen video manipulations and unseen modalities (e.g., event cameras).
Finally, we identify \textit{task internalization}, where a query format associated with a pretrained task overrides the demonstration, and show that this shortcut can be removed with a small amount of task-unrelated data, highlighting the need to decorrelate prompt format from task identity.\footnote{Project page: \url{https://iandrover.github.io/video_analogy/}}
\end{abstract}


\section{Introduction}

Traditionally, cultivating a new capability in visual generative models demands dedicated data curation, architectural modifications, and task-specific fine-tuning. Large language models have demonstrated a compelling alternative: models can learn to execute new tasks simply by observing a few demonstrations in context~\citep{brown2020language,gao2021making,zhao2021calibrate,schick2021s,liu2022makes,dai2023can}. In vision, this principle takes the form of visual analogy~\citep{hertzmann2001image, bar2022visual}, where a model infers a mapping from a source-target image pair and applies it to a novel query. Yet, this capability has remained largely confined to still images, leaving the domain of video largely unexplored.

Extending visual analogy to video is challenging because video tasks demand strict spatiotemporal consistency and complex motion understanding, requirements that currently drive researchers to build specialized datasets and architectures for every new capability. This paper investigates whether a single, analogy-based formulation can absorb this diversity. We ask: \textit{Can a single analogy-based framework effectively unify diverse video tasks and generalize to tasks it has never seen?} 

\begin{figure}[t!]
\centering
\includegraphics[width=0.90\textwidth]{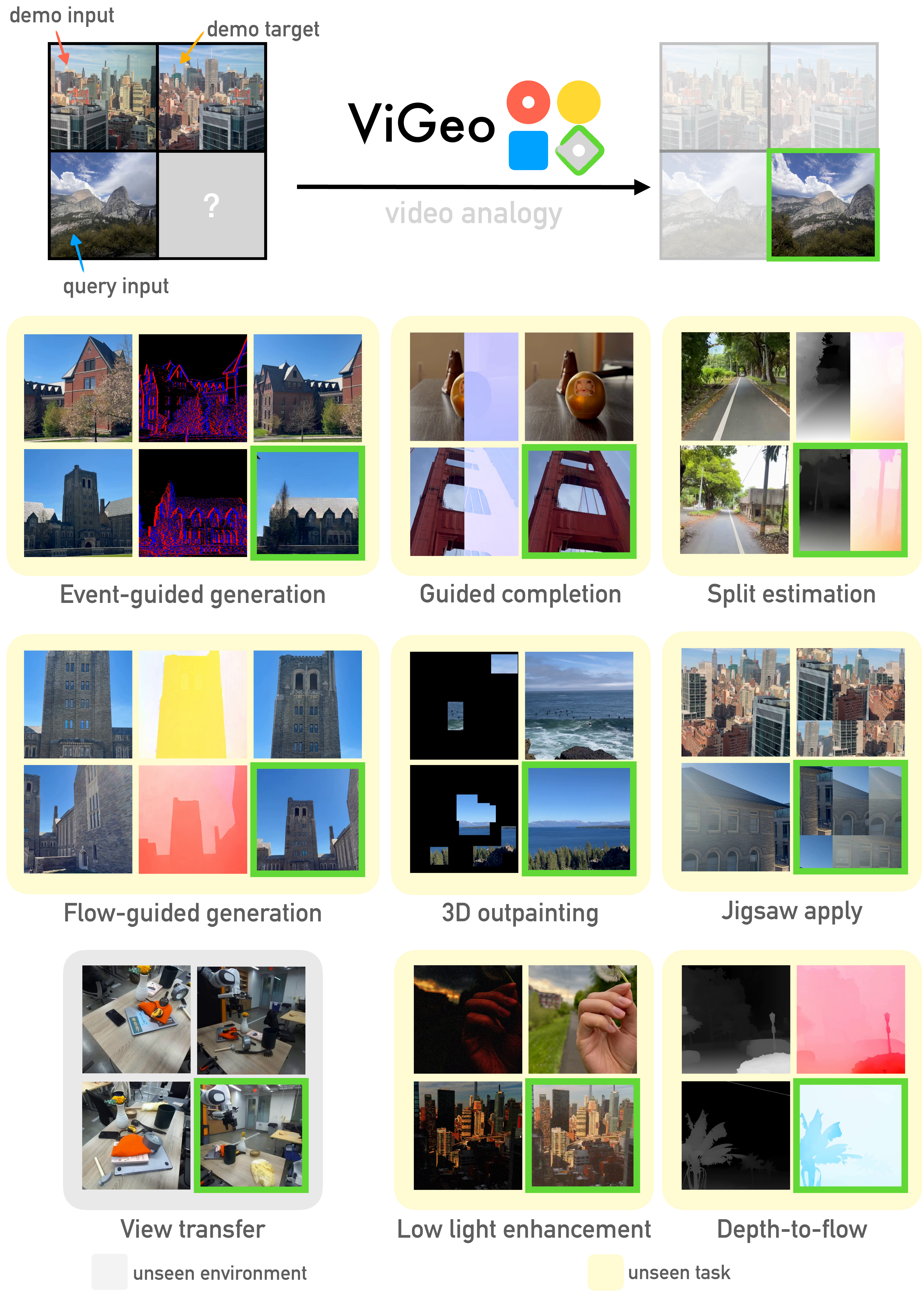}
\caption{
\textbf{\model~unifies and generalizes diverse video tasks via video analogy.} 
\model~achieves \textit{video analogy} by inferring a transformation from a demonstration pair and applying it to a novel query. We show that this formulation unifies a highly diverse taxonomy of tasks, including camera motion and view transfer. Additionally, it exhibits strong generalization to unseen guided generation and video manipulation tasks. \model~predictions are highlighted in green.
\textit{Video results are provided in the \textbf{supplementary material}.}
}
\label{fig:teaser}
\end{figure}

To explore these questions, we must first design a model capable of video analogy without relying on task-specific pipelines. 
We introduce \model
\footnote{Short for \textit{Video, ergo genero}, Latin for ``I see, therefore I generate''---an homage to Descartes.}
, which formulates the problem strictly as a spatiotemporal canvas completion task.
Rather than relying on auxiliary task heads or textual descriptions, \model~arranges the demonstration input, demonstration target, and query input into a single composite video grid. The model is then trained to generate the missing quadrant, i.e., the query target, based purely on the visual context. This framework reduces diverse generative operations to a single, unified visual-completion objective, allowing the model to learn complex mapping relationships entirely from pixels.

While the canvas formulation is task-agnostic in principle, the central question is whether it can support the spatiotemporal complexity of real video operations. 
We therefore evaluate \model~through a structured task taxonomy spanning three categories that demand spatiotemporal understanding: 
camera control, 
modal-guided generation (from signals such as optical flow or point tracks), 
and video manipulation. This taxonomy is designed to test two complementary properties of the framework: whether a single canvas can \textit{unify} diverse in-distribution video tasks, and whether the learned analogy \textit{transfers} to tasks that are never seen during training. Concretely, camera control serves as an in-distribution benchmark against dedicated baselines, while modality-guided generation and video manipulation are evaluated under a strict task-level train-test split. This design allows us to measure genuine in-context learning from the demonstration, rather than memorization of trained task formats. 

Empirically, \model~succeeds on both in-distribution tasks and held-out transfer tasks. On camera control, it remains competitive with task-specific baselines, showing that a single canvas can unify structured in-distribution video operations without task-specific modules. More importantly, it also generalizes beyond its training taxonomy. For modal-guided generation, \model~matches or surpasses purpose-built methods on several reconstruction metrics and transfers to unseen guidance modalities such as event camera data. For video manipulation, it correctly infers and executes the demonstrated transformation on 7 of 12 strictly held-out tasks, indicating that the model is not merely copying inputs but performing in-context learning over video transformations.

Beyond benchmark results, we examine when the model actually relies on its demonstrations. We identify \textit{\textbf{task internalization}}, a failure in which the task is resolved from the query's \emph{format} rather than from the demonstration.
On held-out depth-to-flow task, the same demonstration yields either the desired optical flow or a copy of the RGB video, depending only on whether the RGB hint appears once or is repeated across frames.
We trace this to a format--target shortcut in the pretraining corpus, where every $2\times3$ canvas with a frozen RGB video in query has an RGB target, and show that finetuning on a small amount of data containing neither depth nor optical flow removes the shortcut and restores demonstration-following in both held-out directions.
This suggests that in-context generalization depends not only on which tasks a model is trained on, but on whether the prompt format alone reveals the task.

In summary, our contributions are as follows:
\begin{itemize}
\item \textbf{Formulation:} We propose \model, extending visual analogy to the video domain via a pure spatiotemporal canvas completion framework.
\item \textbf{Rigorous Task-Level Evaluation:} We construct a comprehensive video task taxonomy with a strict task-level train-test split, enabling the explicit measurement of emergent generalization rather than memorization.
\item \textbf{One-Shot Capabilities:} We demonstrate that \model~achieves competitive or even superior performance against task-specific baselines, successfully generalizing to unseen modalities (e.g., event cameras) and novel video manipulation tasks.
\item \textbf{Analysis:} We identify \textit{task internalization}, in which during evaluation a specific query input format overrides the demonstration. We trace it to a format--target shortcut in the training data and remove it with a small amount of task-unrelated data, showing that in-context generalization requires prompt formats that do not themselves reveal the task.
\end{itemize}
\section{Related Work}

\vspace{2mm}
\noindent \textbf{Visual (image) analogies.} 
The idea of transduction with visual demonstration is first explored in Image Analogies~\citep{hertzmann2001image}. 
In this earliest conception, it defines it as ``synthesis of a new filtered target image'' by first inferring the filters from a pair of images in the \textit{design} phase and applying the transformation on a target image in the \textit{application} phase, opening up many earlier works~\citep{reed2015deep, liao2017visual}.

Recently, Visual Prompting~\citep{bar2022visual} fits this idea into a general self-supervised learning scheme by reformulating it as an inpainting/completion pretext task. 
Since then, numerous works have been proposed, such as ~\citep{xu2023improv,
vsubrtova2023diffusion, wang2023images,wang2023context,nguyen2023visual,zhang2024instruct, gu2024analogist,lai2025unleashing,srivastava2025reedit, kim2025difference, li2025visualcloze}.
However, they come with one of the following shortcomings:
(1) Focus on object-centric editing, manipulation, or style transfer tasks~\citep{vsubrtova2023diffusion, nguyen2023visual, srivastava2025reedit, kim2025difference, li2025visualcloze, lai2025unleashing}, limiting their scope to localized or low-level transformation; 
(2) Use auxiliary text prompt~\citep{wang2023context, gu2024analogist, lai2025unleashing}, making it hard to disentangle the effect of the text prompt and image demonstration; or
(3) Evaluate on training tasks, with a few exceptions like ~\citep{li2025visualcloze}. 
In comparison, our \model~operates in the video regime, focus on scene-level, video-prompt setup, and only evaluates on testing tasks or scenarios. 

\noindent \textbf{Concept transfer in video generation.} 
In the video domain, one-shot learning has been focusing on concept transfer in specific domains, such as camera pose~\citep{luo2025camclonemaster, mitchel2025true}, motion~\citep{sun2024video}, and editing~\citep{zhang2025visual}. 
In these works, architectural changes and specific objective functions are designed, but are not yet ready to extend towards diverse video tasks. Our \model~demonstrates a generic framework by drawing visual analogies.
\section{Method}
\label{sec:methodology}

\subsection{Canvas Setup}

Consider a demo input video (\ca $\in \mathbb{R}^{T \times H \times W \times 3}$), a demo target (\cb), a query input (\cc), and a query target (\cd). A video analogy requires the mapping from $A$ to $B$ is the same as the mapping from $C$ to $D$, i.e., $\ca:\cb \sim \cc:\cd$.
We define a canvas construction operator $\mathcal{C}$. This operator spatially arranges the videos into a large composite video canvas $\mathcal{C}(A,B,C,D) \in \mathbb{R}^{T\times (2H+m) \times (2W+m) \times 3}$, forming a $2 \times 2$ grid separated by a margin of size $m$. The demonstrations are placed in the top row and queries in the bottom, yielding $\begin{bsmallmatrix} \ca & \cb \\ \cc & \cd \end{bsmallmatrix}$ (the margin is omitted for clarity). 


\begin{table*}[t]
\centering
\caption{
\textbf{Overview of task taxonomy and train/eval protocol.}
We train on a diverse set of task families that all share a single video analogy canvas $\begin{bsmallmatrix} \ca & \cb \\ \cc & \cd \end{bsmallmatrix}$: a demonstration pair $(\ca, \cb)$ shows the transformation, and the model must generate \cd\ given the query input \cc.
Evaluation tasks are strictly held out to ensure the model never sees them during training.
For $2{\times}3$ layouts, the input spans two grid cells (separated by ``$\mid$'').
See Appendix~\ref{app:task_taxonomy} for full per-task schemas.
}
\label{tab:task_overview}
\small
\renewcommand{\arraystretch}{1.20}
\setlength{\tabcolsep}{4pt}
\begin{tabular}{@{}l l c c c c c@{}}
\toprule
\textbf{Category} & \textbf{Task Family} & $\ca/\cc$ (input) & $\cb/\cd$ (target) & \textbf{Layout} & \textbf{\#} & \textbf{Split} \\
\midrule
\multirow{3}{*}{\textsc{Camera}}
& Motion transfer       & still frame       & video @ traj.       & $2{\times}2$ & 1 & Train + Eval \\
& View transfer         & video @ pose 1    & video @ pose 2      & $2{\times}2$ & 2 & Train + Eval \\
\midrule
\multirow{8}{*}{\textsc{Modal}}
& Modality estimation   & rgb video         & modality video      & $2{\times}2$ & 5 & Train \\
& Multi-modal quad      & rgb video         & 4-modal grid        & $2{\times}2$ & 1 & Train \\
& Segmentation          & rgb video   & mask / matted       & $2{\times}\{2,3\}$ & 2 & Train \\
& Guided generation     & frozen rgb vid. $\mid$ modal & rgb video      & $2{\times}3$ & 4 & Train \\
\cmidrule(l){2-7}
& \emph{Flow-guided gen}   & frozen rgb vid. $\mid$ flow viz   & rgb video & $2{\times}3$ & 1 & Eval \\
& \emph{Point-guided gen}  & frozen rgb vid. $\mid$ point map  & rgb video & $2{\times}3$ & 1 & Eval \\
& \emph{Event-guided gen}  & 2 anchor frames $\mid$ event viz & rgb video & $2{\times}3$ & 1 & Eval \\
\midrule
\multirow{8}{*}{\textsc{Manip.}}
& Zoom (in/out)         & wide view         & close-up            & $2{\times}2$ & 4 & Train \\
& Photometric editing   & original          & edited              & $2{\times}2$ & 1 & Train \\
& Spatiotemporal perm.  & permuted          & original            & $2{\times}2$ & 4 & Train \\
\cmidrule(l){2-7}
& \emph{Inpainting / outpainting} & masked video & original    & $2{\times}2$ & 2 & Eval \\
& \emph{Deblurring}     & blurred           & sharp               & $2{\times}2$ & 3 & Eval \\
& \emph{Temporal interp.} & sparse frames   & full video          & $2{\times}2$ & 2 & Eval \\
& \emph{Jigsaw (2D)}    & shuffled tiles    & original            & $2{\times}2$ & 2 & Eval \\
& \emph{Symmetry / low-light} & flipped or dark & original       & $2{\times}2$ & 3 & Eval \\
\bottomrule
\end{tabular}
\end{table*}

\subsection{Task Design}
\label{sec:task_design}

As the canvas treats each block as a pixel video,
a wide range of video tasks fit the same interface:
RGB scenes, depth maps, flow visualizations, segmentation overlays, and point trajectories all occupy the same pixel space,
enabling a single model to absorb diverse tasks.
We exploit this flexibility to unify a diverse set of video tasks under a single framework, spanning three categories: camera control, modal-guided generation, and video manipulation.
Table~\ref{tab:task_overview} summarizes the full taxonomy.

\noindent \textbf{Training tasks.}
We train on over 20 task families drawn from complementary data sources, including SpatialVID~\citep{wang2025spatialvid}, DL3DV~\citep{ling2024dl3dv}, SynCamVideo~\citep{bai2024syncammaster}, CameraClone~\citep{luo2025camclonemaster} BridgeData-v2~\citep{walke2023bridgedata}, DROID~\citep{khazatsky2024droid}, 
SA-V~\citep{kirillov2023segment}, 
and Hypersim~\citep{roberts2021hypersim}.
These tasks collectively expose the model to modality estimation and generation, geometric camera transformations, and spatiotemporal permutations, all presented through the same canvas interface without any task-specific text. 

\noindent \textbf{Evaluation protocol.}
To rigorously test generalization, we enforce a strict task-level train/test split: evaluation tasks are never seen during training.
The held-out tasks probe three transfer axes, including
(1)~\emph{unseen conditioning modality}: flow, point, and event-camera guidances are never used as conditioning signals during training, and event-camera is the genuine unseen-modality case as it is absent from every training task;
(2)~\emph{unseen operation}: 2D jigsaw, deblurring, and flipping are structurally distinct from the 1D strip shuffles, warps, and photometric edits seen during training;
(3)~\emph{unseen environment}: camera view transfer tasks are evaluated on physically held-out buildings and rooms.
This design ensures that strong evaluation performance reflects genuine in-context performance from the demonstration.
Furthermore, we split the source videos to train set and test set, so that any clip used in the test time, whether as a source or target and whether in the demo or query row, is not seen during training.

\subsection{Learning Objective}
\label{subsec:learning_objective}

We train the model to complete the target quadrant \cd\ using a Rectified Flows objective~\citep{liu2022flow}.
Let $\mathcal{E}$ denote the video encoder and $f$ be our generative model.
During training, our target data point is the encoded query target, $x_1 = \mathcal{E}(D)$, and we sample standard Gaussian noise $x_0 \sim \mathcal{N}(0, I)$. For a given time step $t \in [0,1]$, the noisy latent $x_t$ is defined as a linear interpolation between the noise and the target: $x_t = t x_1 + (1-t) x_0$, whereas the ground-truth velocity $v_t$ is defined as
$v_t = \frac{dx_t}{dt} = x_1 - x_0$.
The model is trained to predict this velocity, and the loss function is formulated as the mean squared error between the model's prediction and the true velocity.

To condition the prediction on the canvas context, we make a few adjustments. Let $M$ denote a zero-mask video of the same dimensions as $D$. We first encode the masked canvas, obtaining the contextual latents $\mathcal{E}(\mathcal{C}(A, B, C, M))$. We then construct the full noisy canvas latent, denoted as $z_t$, by replacing the quadrant corresponding to the masked query target $M$ with our noisy state $x_t$. The generative model $f$ takes this composite latent $z_t$ and the timestep $t$ as inputs to predict the velocity of the target region. The overall training objective is defined as:
\begin{equation}
    \mathcal{L} = \mathbb{E}_{x_0, x_1, t} \left[ \left\| v_t - f(z_t, t) \right\|_2^2 \right]
\end{equation}

\subsection{Training Setup}
All the videos (i.e., \ca, \cb, \cc, \cd) are cropped and resized to $480\times 480$, with a length of $81$. With a margin $m$ of $32$, the canvas spatial resolution is $992\times 992$ or  $992\times 1504$, corresponding to
$2\times 2$ or $2\times 3$ layout.
We perform full finetuning on Wan2.2-TI2V-5B~\citep{wan2025} with a learning rate of $0.00005$ and a batch size of $512$ for 10K steps.
The video encoder has a pixel patch size of $16\times 16\times 4$ and a latent patch size of $2 \times 2 \times 1$, and thus we select the margin size $m$ to be $32$.
We use FastVideo~\citep{zhang2025fast} and train the model across $32$ NVIDIA A100 nodes.

\section{Experiments}
\label{sec:experiments}

\subsection{Modal Guided Generation}

\begin{table}[t]
\centering
\caption{Quantitative comparison of our method against strong task-specific baselines on flow-, point-, and event-guided generation. Notably, all three tasks are unseen to our model. The \textbf{best} and \underline{second best} results are bolded and underlined, respectively.}
\label{tab:guided_generation}
\small
\setlength{\tabcolsep}{8pt} 
\begin{tabular}{llc ccc}
\toprule
& & \textbf{Optical Flow} & \multicolumn{3}{c}{\textbf{Reconstruction Metrics}} \\
\cmidrule(lr){3-3} \cmidrule(lr){4-6}
\textbf{Task} & \textbf{Method} & \textbf{EPE $\downarrow$} & \textbf{PSNR $\uparrow$} & \textbf{SSIM $\uparrow$} & \textbf{LPIPS $\downarrow$} \\
\midrule

\multirow{3}{*}{\begin{tabular}[c]{@{}l@{}}\task{Flow-}Guided \\[-0.5ex] 
Generation
\end{tabular}} 
& FloVD \tiny{(CVPR 2025)} & 3.771 & 12.363 & 0.240 & 0.539 \\
& Go-with-the-flow \tiny{(CVPR 2025)} & \underline{0.356} & \underline{19.214} & \underline{0.491} & \underline{0.181} \\
& \cellcolor{lightblue}\model & \cellcolor{lightblue}\textbf{0.185} & \cellcolor{lightblue}\textbf{22.365} & \cellcolor{lightblue}\textbf{0.671} & \cellcolor{lightblue}\textbf{0.106} \\
\midrule

\multirow{3}{*}{\begin{tabular}[c]{@{}l@{}}\task{Point-}Guided \\[-0.5ex] 
Generation
\end{tabular}} 
& Tora \tiny{(CVPR 2025)} 
& 0.612 
& \phantom{0}6.923 
& \underline{0.366} 
& 0.230 \\
& Wan-Move \tiny{(NeurIPS 2025)} 
& \underline{0.289} 
& \underline{17.048} 
& 0.365 
& \textbf{0.157} \\
& \cellcolor{lightblue}\model 
& \cellcolor{lightblue}\textbf{0.275} 
& \cellcolor{lightblue}\textbf{19.515} 
& \cellcolor{lightblue}\textbf{0.543} 
& \cellcolor{lightblue}\underline{0.226} \\
\midrule

\multirow{3}{*}{\begin{tabular}[c]{@{}l@{}}\task{Event-}Based \\[-0.5ex] 
Interpolation
\end{tabular}} 
& CBMNet \tiny{(CVPR 2023)} & 8.020 & \textbf{23.587} & \textbf{0.759} & 0.284 \\
& RE-VDM \tiny{(CVPR 2025)} & \textbf{3.384} & 20.892 & \underline{0.665} & \underline{0.239} \\
& \cellcolor{lightblue}\model 
& \cellcolor{lightblue}\underline{4.539} 
& \cellcolor{lightblue}\underline{21.035} 
& \cellcolor{lightblue}{0.652} 
& \cellcolor{lightblue}\textbf{0.195} \\
\bottomrule
\end{tabular}
\end{table}

\noindent \textbf{Setup.}
We evaluate three guided generation tasks, all strictly held out during training.
Each task uses a $2{\times}3$ canvas layout.
Specifically, for flow- and point-guided generation, the input (\ca/\cc) consists of the first frame alongside the modality visualization, whereas the target is the full RGB video (\cb/\cd).
For event-based interpolation, two anchor RGB frames are provided and the target is the full RGB video. 
The three evaluation modalities are optical flow, point trajectories (16 uniformly sampled points), and event camera data.
We compare against task-specific baselines: FloVD~\citep{jin2025flovd} and Go-with-the-flow~\citep{burgert2025go} for flow; Tora~\citep{zhang2025tora} and Wan-Move~\citep{chu2026wan} for points; CBMNet~\citep{kim2023event} and RE-VDM~\citep{chen2025repurposing} for event-based interpolation, where our model digests rendered event data visualization.
Since these tasks consider dense control signals, we adopt metrics including optical flow end-point error (EPE) and reconstruction quality (PSNR, SSIM, LPIPS).

Table~\ref{tab:guided_generation} presents the quantitative results. For flow-guided generation, \model~achieves exceptional performance across all metrics. Despite lacking any prior exposure to optical flow as conditioning signals, it yields the lowest end-point error (EPE) and the highest reconstruction scores (PSNR, SSIM, and LPIPS), outperforming even dedicated flow-conditioned architectures.

In the point-guided setting, conditioned on sixteen, uniform points, \model~remains highly competitive against purpose-built baselines such as Tora~\citep{zhang2025tora} and Wan-Move~\citep{chu2026wan}. It secures the lowest optical flow EPE and the highest PSNR and SSIM scores, demonstrating its robust ability to parse and extrapolate from spatial prompts.

Finally, we push the model's out-of-distribution generalization further by evaluating video frame interpolation guided by event camera data, a visual representation entirely absent from the training distribution. \model~adapts to this novel sensory input as rendered images and achieves the best LPIPS score among all methods. The remaining metrics reveal a trade-off relative to the specialized baselines: \model~achieves lower optical flow EPE than CBMNet~\citep{kim2023event} and higher PSNR than RE-VDM~\citep{chen2025repurposing}, but it trails RE-VDM in EPE and CBMNet in both PSNR and SSIM. We attribute this gap to the fact that both baselines are designed and trained specifically for event-based interpolation, whereas \model~handles event inputs without task-specific design. These results suggest that \model~produces perceptually faithful interpolations on an unseen modality, while specialized methods retain an advantage in pixel-level fidelity and motion accuracy.

\subsection{Camera Control}
\label{subsec:exp_camera_control}

\begin{table}[thbp]
\centering
\caption{Quantitative comparison of our method on the \task{camera\_motion\_transfer} task. The \textbf{best} and \underline{second best} results are bolded and underlined, respectively.}
\label{tab:camera_motion_transfer}
\small
\setlength{\tabcolsep}{5pt} 
\begin{tabular}{lll ccc}
\toprule
& & & \multicolumn{3}{c}{\textbf{Camera Metrics}} \\
\cmidrule(lr){4-6}
\textbf{Task} & \textbf{Dataset} & \textbf{Method} & \textbf{RotErr $\downarrow$} & \textbf{TransErr $\downarrow$} & \textbf{CamErr $\downarrow$} \\
\midrule

\multirow{3}{*}{\begin{tabular}[c]{@{}l@{}}Camera \\[-0.5ex] \task{Motion} \\
[-0.5ex] Transfer
\end{tabular}}
& \multirow{3}{*}{DL3DV} 
& CamCloneMaster \tiny{(SIGGRAPH Asia 2025)} 
& \phantom{}3.367 
& \underline{13.941}
& 15.408 \\
& & XFactor \tiny{(ICLR 2026)} 
& \textbf{\phantom{}1.066} 
& \textbf{\phantom{0}5.378} 
& \textbf{\phantom{0}5.761} \\
& & \cellcolor{lightblue}\model 
& \cellcolor{lightblue}\phantom{}\underline{2.173} 
& \cellcolor{lightblue}{14.612}
& \cellcolor{lightblue}\underline{15.203} \\

\bottomrule
\end{tabular}
\end{table}

\begin{table}[thbp]
\centering
\caption{Quantitative result on the \task{camera\_view\_transfer} task. The \textbf{best} results are bolded.}
\label{tab:camera_view_transfer}
\small
\setlength{\tabcolsep}{5pt} 
\begin{tabular}{lll ccc}
\toprule
& & & \multicolumn{3}{c}{\textbf{Camera Metrics}} \\
\cmidrule(lr){4-6}
\textbf{Task} & \textbf{Dataset} & \textbf{Method} & \textbf{RotErr $\downarrow$} & \textbf{TransErr $\downarrow$} & \textbf{CamErr $\downarrow$} \\
\midrule

\multirow{4}{*}{\begin{tabular}[c]{@{}l@{}}Camera \\[-0.5ex] \task{View} \\
[-0.5ex] Transfer
\end{tabular}}
& \multirow{2}{*}{BridgeData-v2} 
& XFactor \tiny{(ICLR 2026)} 
& 36.131
& 0.857
& 1.235 \\
& & \cellcolor{lightblue}\model & \cellcolor{lightblue}\textbf{\phantom{}2.382} & \cellcolor{lightblue}\textbf{0.093} & \cellcolor{lightblue}\textbf{0.117} \\
\cmidrule(lr){2-6}
& \multirow{2}{*}{DROID} 
& XFactor \tiny{(ICLR 2026)} 
& 64.824
& 1.200 
& 1.959 \\
& & \cellcolor{lightblue}\model & \cellcolor{lightblue}\textbf{\phantom{}9.149} & \cellcolor{lightblue}\textbf{0.116} & \cellcolor{lightblue}\textbf{0.261} \\
\bottomrule
\end{tabular}
\end{table}

\noindent \textbf{Setup.}
We evaluate two camera control tasks using a $2{\times}2$ canvas.
For \task{camera\_motion\_transfer}, the demonstration shows a static frame (\ca) animated along a trajectory (\cb); the model must apply the same camera motion to the query static frame (\cc). We evaluate on DL3DV~\citep{ling2024dl3dv}.
For \task{camera\_view\_transfer}, the demonstration shows a scene from one viewpoint (\ca) captured at a different pose (\cb, with synchronized cameras); the model generates \cd\ at the target pose given query input \cc. We evaluate on held-out environments from BridgeData-v2~\citep{walke2023bridgedata} and DROID~\citep{khazatsky2024droid}.
Baselines include XFactor~\citep{mitchel2025true} (frame-level) and CamCloneMaster~\citep{luo2025camclonemaster} (video-level).
Metrics are rotation error (RotErr, in degrees), translation direction error (TransErr), and combined camera error (CamErr); see Appendix~\ref{subsec:exp_camera_view_transfer_full} for reconstruction and covisible metrics.

In Table~\ref{tab:camera_motion_transfer}, we benchmark \model~against both frame-level and video-level baselines for the \task{camera\_motion\_transfer} task. While XFactor achieves the highest overall scores on this specific task, \model~achieves highly competitive performance and outperforms CamCloneMaster on both rotation and overall camera error.

For \task{camera\_view\_transfer}, \model~achieves lower camera errors than XFactor on held-out environments from BridgeData-v2 and DROID (Table~\ref{tab:camera_view_transfer}). We note that this comparison may be unfair due to a difference in training domains: \model's training data includes such dual-camera setup from both datasets, whereas XFactor is trained on human-captured videos. As shown in our analysis in Appendix~\ref{subsec:xfactor_failure_analysis}, XFactor's degradation may stem from the large roll and pitch angles of the dual-camera setup, which are rare in its training data. Taken together, our results therefore suggest that, when trained on diverse camera distributions, the video analogy formulation allows \model~to infer large multi-axis rotations from a single demonstration pair and transfer them to a new query.

\subsection{Video Manipulation}
\label{subsec:experients_video_manipulation}

\begin{figure*}[t]
\centering
\includegraphics[width=\textwidth]{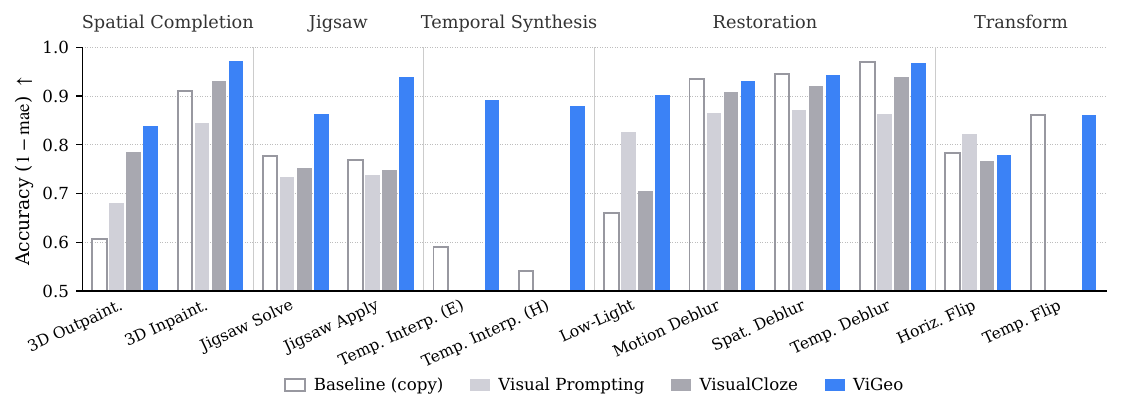}
\caption{
\textbf{Generalization to unseen video processing tasks.}
We evaluate on 12 held-out video processing tasks and report the accuracy between the model output and the target video.
Our model generalizes to 7 of 12 unseen tasks, while deblurring and geometric flipping remain challenging, and the model merely copies the input query video.
}
\label{fig:exp_video_processing}
\end{figure*}

\noindent \textbf{Setup.}
We evaluate \model~on 12 unseen tasks using a $2{\times}2$ canvas.
These tasks span spatiotemporal completion (2), jigsaw (2), temporal synthesis (2), restoration (4), and spatiotemporal transforms (2).
We compare against a naive copy-input baseline, Visual Prompting~\citep{bar2022visual}, and VisualCloze~\citep{li2025visualcloze}, a recent model demonstrating emergent out-of-domain capabilities.
We calculate the pixel-wise mean absolute error (MAE) and report accuracy as $1 - \text{MAE}$.

As shown in Figure~\ref{fig:exp_video_processing}, on 7 of the 12 tasks, the output of \model~is substantially better than all three baselines, confirming that it has successfully inferred and applied the demonstrated transformation rather than defaulting to a copy. The model handles spatial permutations well, effectively applying and inverting jigsaw shuffles, and performs reliable temporal interpolation. For strongly generative tasks such as 3D outpainting, while the absolute accuracy is comparatively lower than inpainting, it is likely because it requires hallucinating novel content rather than rearranging existing pixels, \model~still demonstrates a significant improvement over the copy baseline.

Conversely, the model struggles with the remaining 5 tasks (deblurring and flipping), where it fails to generalize and defaults to the copy baseline. This divergence in performance suggests a consistent pattern: operations that preserve local pixel structure transfer more successfully, while those requiring global geometric reorganization (flipping) or the inversion of a destructive process (deblurring) do not. We hypothesize that test-time generalization to unseen tasks is not unconditional; rather, it requires the underlying transformation to share structural primitives with those seen during training. Because deblurring and flipping involve spatial convolution and global permutation operations that are absent from our training taxonomy, the model struggles to infer them. This implies that for future work, increasing the structural diversity of the training taxonomy may be more critical than simply expanding the dataset size for existing tasks.
\section{A Deep Dive to Video Analogy}
\label{sec:deep_dive}

The same analogy can be expressed through different query formulations. 
For example, an unseen depth-to-flow translation can be prompted with a [depth; optical flow] demo alone. 
But what happens if we add an RGB hint in the query like [first-frame RGB, depth; optical flow]? 
In this section, we study two sensitivities: (1) \textit{task internalization} (Section~\ref{subsec:task_internalization}), where a familiar query format overrides the demonstration and the model executes the trained task instead of the demonstrated analogy. We then break this format--task coupling by finetuning with an additional normal-to-edge task to mitigate the problem. (2) \textit{perceptual invariance} (Appendix~\ref{subsec:perceptual_invariance}), where the model abstractly interprets modal guidance across different color palettes.

\begin{figure*}[t]
\centering
\includegraphics[width=\textwidth]{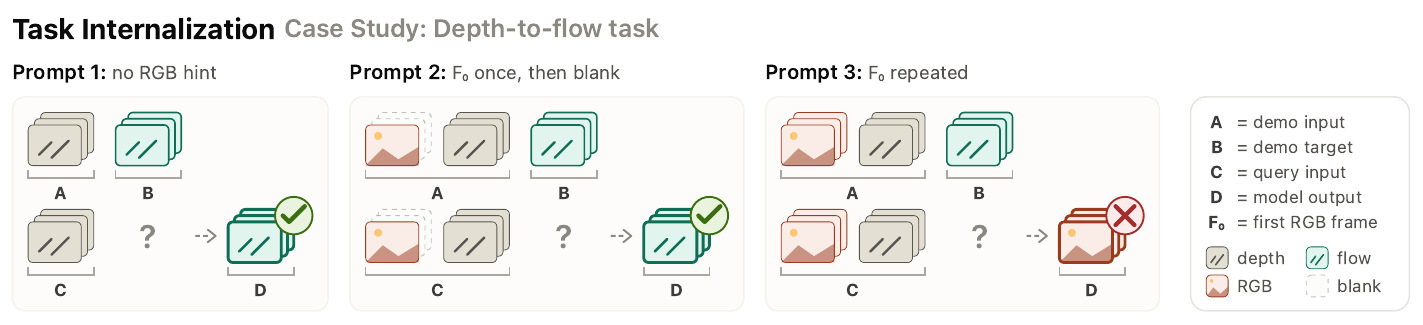}
\caption{\textbf{Illustration of \textit{task internalization}.}
We test an unseen depth-to-flow translation task under three different demo prompts. In prompt 1 (no RGB image provided, $2\times2$ format) and prompt 2 (RGB hint provided, where the first RGB frame appears once), the model succeeds in outputting optical flow visualization video. In prompt 3, where the RGB hint is the first frame repeated across all frames, the model's outputs collapse to RGB videos. 
}
\label{fig:task_intn_teaser}
\end{figure*}

\subsection{Task Internalization: The Interplay Between Familiar Queries and Novel Demonstrations}
\label{subsec:task_internalization}

\noindent \textbf{Probing Setup.}
As shown in Figure~\ref{fig:task_intn_teaser}, we compare three query formats: (1)~depth only, in the $2\times2$ format; (2)~the first RGB frame followed by black frames, in the $2\times3$ format; and (3)~the first RGB frame repeated across all frames, i.e.\ a frozen video. The task is out-of-distribution, so a well-behaved model should follow the input--target relation given in the demonstration and output an optical-flow visualization in all three cases. 
To detect the failure mode, we measure the Pearson correlation between the generated block and the query clip's RGB video; a high value means the model reproduced the RGB video instead of the demanded modality.
We run both depth-to-flow and flow-to-depth, with $n{=}55$ samples each task.

\begin{figure*}[t]
\centering
\includegraphics[width=\textwidth]{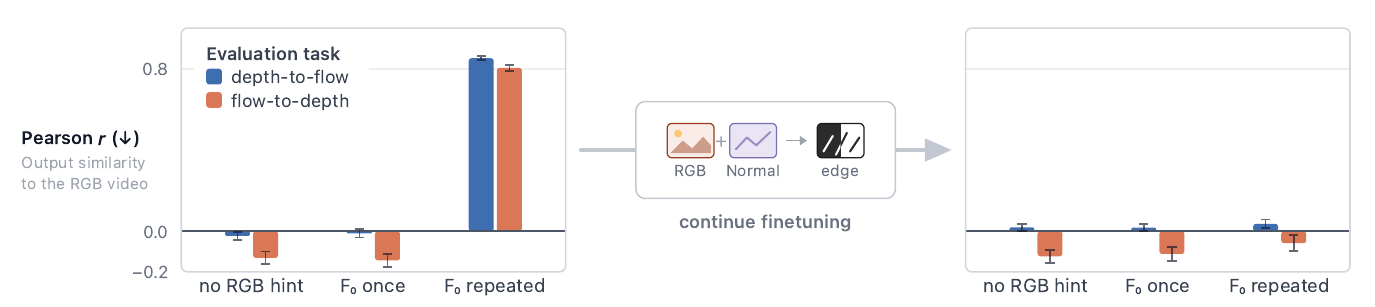}
\caption{\textbf{Task internalization, and its mitigation.}
Correlation between the generated output and the query's RGB video is computed; a higher value means the model outputs RGB instead of the demanded modality.
\textbf{Left:} the pretrained model spikes to $0.85$ and $0.81$ under the repeated RGB hint, in both held-out tasks, and sits near zero in every other condition.
\textbf{Right:} after finetuning on a normal-to-edge task that contain neither depth nor optical flow, the spike disappears.}
\label{fig:task_intn_exp}
\end{figure*}

\noindent \textbf{Findings.}
The left panel of Figure~\ref{fig:task_intn_exp} shows the result. Under prompt~3 the model outputs RGB videos, reaching correlations of $0.85$ (depth-to-flow) and $0.81$ (flow-to-depth) with the RGB input. Under prompts~1 and~2 the same model produces the correct
modality, with correlations near zero (e.g., $-0.02$ and $-0.01$ respectively for the depth-to-flow task). The results have two implications. First, the failure is \emph{format-dependent rather than task-dependent}: the outcome flips according to how the RGB hint is
given. Second, \emph{a single RGB frame does not trigger it} --- prompt~2 also supplies RGB, yet behaves like the no-hint condition. 

\noindent\textbf{Hypothesis: a format--target shortcut.}
Upon inspecting the training task composition, we find a structural pattern that potentially accounts for this phenomenon. 
In the training tasks, every 2×3 canvas containing a frozen RGB video query input has an RGB video as its target.
The implication of \emph{a $2\times3$ layout with a frozen RGB column $\Rightarrow$ RGB target} therefore holds. 
We term the behavior \emph{\textbf{task internalization}}: once the query matches a familiar task template, the model collapses to its pretrained prior. 
The pattern mirrors \emph{cue competition}~\citep{rescorla1972theory}, in which two informative cues are presented together and the more salient one absorbs the prediction error. 
This also echoes the in-context learning behaviors in large language models: when the query format suffices to recognize a pretrained task, the demonstration is ignored~\citep{min2022rethinking}, and genuine task learning occurs only when recognition fails~\citep{pan2023context}.

\subsection{Mitigating Task Internalization}
\label{subsec:mitigation}

\noindent\textbf{Insight.}
Since the shortcut is structural rather than task-specific, simply training on the tasks of interest would invalidate the held-out split. Yet, this same property itself suggests a solution: because the shortcut is not tied to any task, we may be able to break the format–target correlation using unrelated content.
We can construct a $2\times 3$-layout task where the target are not RGB videos anymore, forcing the model to check demo targets, and break the format--target shortcut.

\noindent\textbf{Experiment.}
We build a small-scale normal-to-edge transformation dataset (1.1\% of the original training corpus) and mix it into the original pool with a $10\times$ oversample and continue finetuning the model for $750$ steps.
The right panel of Figure~\ref{fig:task_intn_exp} shows the effect. Under the repeated RGB hint prompt, correlation with the RGB video falls from $0.85$ to $0.04$ for depth-to-flow and from $0.81$ to $-0.06$ for flow-to-depth. See Section~\ref{subsec:task_int_exp_flow_and_depth} for more finetuned model results.
\section{Conclusion}

We introduced \model, a framework that extends visual analogy to the video domain by reducing diverse video tasks to a single spatiotemporal canvas completion objective.
By arranging demonstration and query videos into a unified grid and training the model to generate the missing target quadrant, \model~absorbs a broad taxonomy of video operations, including camera control, modal-guided generation, and video manipulation, without task-specific text, architectures, or modules.
Evaluated under a strict task-level train-test split, \model~matches or is competitive with dedicated baselines on entirely unseen modalities (event cameras) and successfully generalizes to $7$ of $12$ held-out video manipulation tasks.
Finally, our discovery of task internalization, where familiar query formats can override in-context demonstrations, provides critical insights into the design principles of spatiotemporal prompting for future generalist video models.



\newpage
\bibliographystyle{plainnat}
\bibliography{reference}


\newpage
\appendix
\section{More Experiments}

\subsection{Visualization}
\begin{figure}[t!]
\centering
\includegraphics[width=\textwidth]{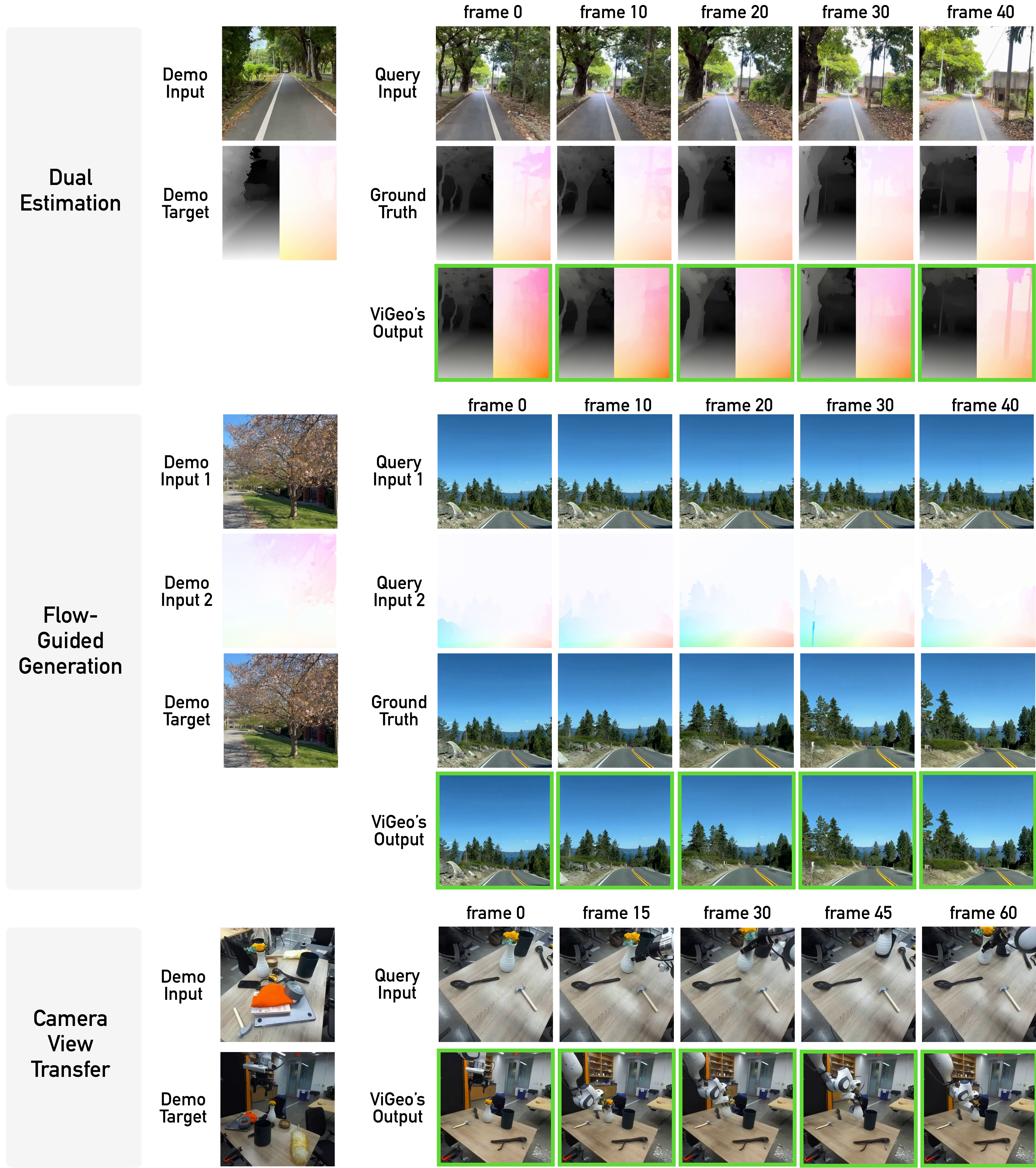}
\caption{
\textbf{Qualitative results across diverse video tasks.} 
We visualize the model's predictions for depth-flow dual estimation, flow-guided generation, and camera view transfer. 
For the demonstration videos in the context, we display only the initial frame ($t=0$) for brevity, while the query and generated target quadrants are shown as temporal rollouts. 
The results highlight \model's ability to synthesize high-fidelity, motion-consistent video while adhering to the geometric and semantic constraints provided by the in-context examples.
}
\label{fig:more_restuls}
\end{figure}

We provide qualitative rollouts generated by \model~across several challenging scenarios: depth-flow dual estimation, flow-guided video generation, and camera view transfer in Figure~\ref{fig:more_restuls}. These visualizations demonstrate the model's ability to maintain spatiotemporal consistency across diverse modalities on a single unified canvas.

\subsection{Camera Control}
\label{subsec:exp_camera_view_transfer_full}

In addition to Table~\ref{tab:camera_view_transfer}, we provide a more detailed comparison between XFactor~\citep{mitchel2025true} and \model~in Table~\ref{tab:camera_view_transfer_reconstruction}. Here, we consider Reconstruction Metrics along with the Covisible Metrics, which calculate the metrics on pixels that are co-visible (estimated using VGGT~\citep{wang2025vggt}) from both poses of the query input and query target. In both cases, \model~obtains superior performance. 

\begin{table}[h]
\centering
\caption{Quantitative comparison of our method on camera view transfer. The \textbf{best} results are bolded.}
\label{tab:camera_view_transfer_reconstruction}
\small
\setlength{\tabcolsep}{5pt} 
\begin{tabular}{lll ccc cc}
\toprule
& & & \multicolumn{3}{c}{\textbf{Full Frame}} & \multicolumn{2}{c}{\textbf{Covisible}} \\
\cmidrule(lr){4-6} \cmidrule(lr){7-8}
\textbf{Task} & \textbf{Dataset} & \textbf{Method} & \textbf{PSNR $\uparrow$} & \textbf{SSIM $\uparrow$} & \textbf{LPIPS $\downarrow$} & \textbf{PSNR $\uparrow$} & \textbf{SSIM $\uparrow$} \\
\midrule

\multirow{4}{*}{\begin{tabular}[c]{@{}l@{}}Camera \\[-0.5ex] \task{View} \\[-0.5ex] Transfer\end{tabular}}

& \multirow{2}{*}{BRIDGE} 
& XFactor \tiny{(ICLR 2026)} 
& 13.268 
& 0.502 
& 0.604 
& 14.279 
& 0.524 \\
& & \cellcolor{lightblue}\model 
& \cellcolor{lightblue}\textbf{16.773} 
& \cellcolor{lightblue}\textbf{0.622} 
& \cellcolor{lightblue}\textbf{0.299} 
& \cellcolor{lightblue}\textbf{16.908} 
& \cellcolor{lightblue}\textbf{0.608} \\

\cmidrule(lr){2-8}

& \multirow{2}{*}{DROID} 
& XFactor \tiny{(ICLR 2026)} 
& 12.047 
& 0.480 
& 0.666 
& 12.186 
& 0.498 \\
& & \cellcolor{lightblue}\model 
& \cellcolor{lightblue}\textbf{14.470} 
& \cellcolor{lightblue}\textbf{0.564} 
& \cellcolor{lightblue}\textbf{0.377} 
& \cellcolor{lightblue}\textbf{14.416} 
& \cellcolor{lightblue}\textbf{0.569} \\
\bottomrule
\end{tabular}
\end{table}

\begin{table}[thb]
\centering
\caption{Quantitative comparison of \task{camera\_view\_transfer} where the demo and query videos are from different hold-out environments.}
\label{tab:camera_view_transfer_hard}
\footnotesize
\setlength{\tabcolsep}{6.0pt}
\begin{tabular}{l cc cc}
\toprule
& \multicolumn{2}{c}{\textbf{DROID ($N=99$)}} & \multicolumn{2}{c}{\textbf{BRIDGE ($N=100$)}} \\
\cmidrule(lr){2-3} \cmidrule(lr){4-5}
\textbf{Method} & \textbf{RotErr} $\downarrow$ & \textbf{TransDir} $\downarrow$ & \textbf{RotErr} $\downarrow$ & \textbf{TransDir} $\downarrow$ \\
\midrule
XFactor \tiny{(ICLR 2026)} 
& 68.92 & 63.64 
& 39.48 & 49.12 \\
\rowcolor{lightblue}
\model
& \textbf{31.58} & \textbf{18.77} 
& \textbf{27.32} & \textbf{22.79} \\
\bottomrule
\end{tabular}
\end{table}

During training, \model~leverages the environment context from both the demo and query, and this yields high reconstruction performance (Table~\ref{tab:camera_view_transfer}). A natural next question is to challenge \textit{what would happen if the demo and query videos are from different environments?} This means that the model can no longer leverage the information from the demo environment and has to extract exclusively the camera pose from the demo pair and the environment context from the query input. We construct this experiment by cross-pairing the demos and queries from different environments. As shown in Table~\ref{tab:camera_view_transfer_hard}, our performance worsens compared to the more aligned scenario. 

However, we argue that this failure mode is expected and comes naturally with the current task setup. During training time, we do not provide training scenarios where the environment of the demo videos and query videos is not the same, causing the model to build a shortcut towards using the context from both query and demo rows. A future next step is to construct synthetic demo videos that share the same relative pose difference as the query videos to encourage better pose and context disentanglement.

\subsection{Supplementary Experiment on Task Internalization: Flow $\rightarrow$ Depth Video Generation}

Similar to the depth-to-flow experiment in Section~\ref{subsec:task_internalization}, we show the result on the held-out flow-to-depth cross-modal video translation task, where a similar task internalization pattern is observed.

\begin{figure}[th]
\centering
\includegraphics[width=\textwidth]{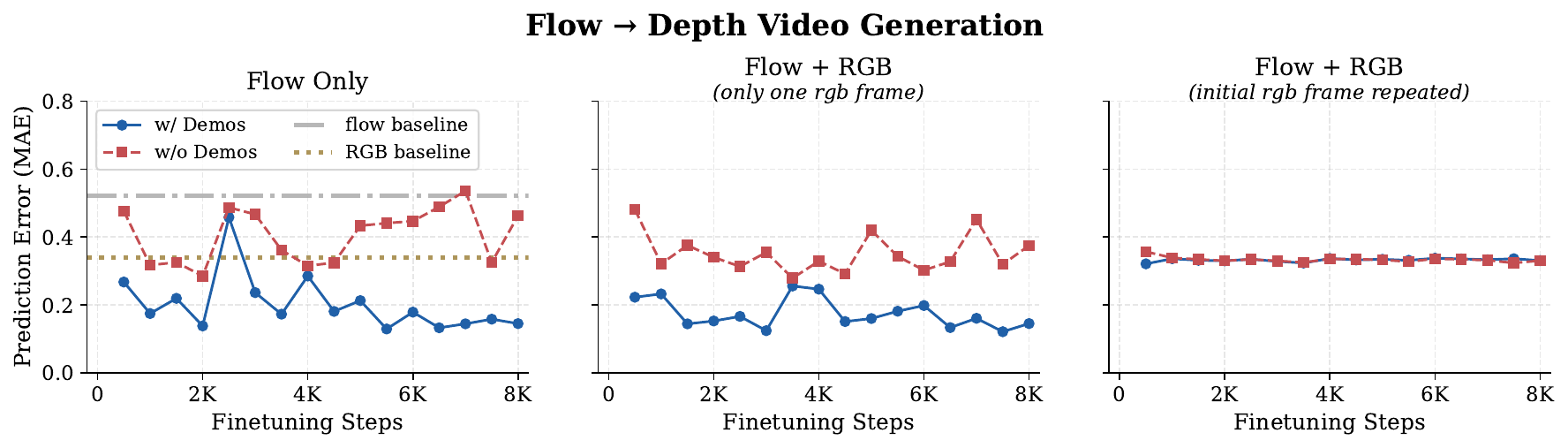}
\vspace{-6mm}
\caption{
\textbf{Query formats that resemble training inputs trigger task internalization.} 
}
\label{fig:icl_prompting_flow_to_depth}
\end{figure}

\subsection{Failure Mode Analysis of XFactor}
\label{subsec:xfactor_failure_analysis}

The severe performance drop of XFactor~\citep{mitchel2025true} on the \task{camera\_view\_transfer} task (Section~\ref{subsec:exp_camera_control}) warrants further investigation. To understand why a model trained on extensive camera trajectory data fails in robotic environments, we conduct a qualitative failure mode analysis.

We hypothesize that the extreme camera rotations inherent to robotic manipulation represent a critical out-of-distribution (OOD) shift for XFactor. To test this, we evaluate the model across four demo-query pair combinations using two distinct domains: a general outdoor scene (representative of standard trajectory datasets like DL3DV) and an indoor robotic manipulation setup (captured via left and right side cameras observing a Franka robot, a common configuration in the DROID dataset~\citep{khazatsky2024droid}). The task requires XFactor to observe the relative camera transformation in the demo pair (demo input to demo target) and apply that exact transformation to the query input.

\begin{figure}[htbp]
    \centering
    
    \begin{subfigure}[b]{0.35\textwidth}
        \centering
        \includegraphics[width=\textwidth]{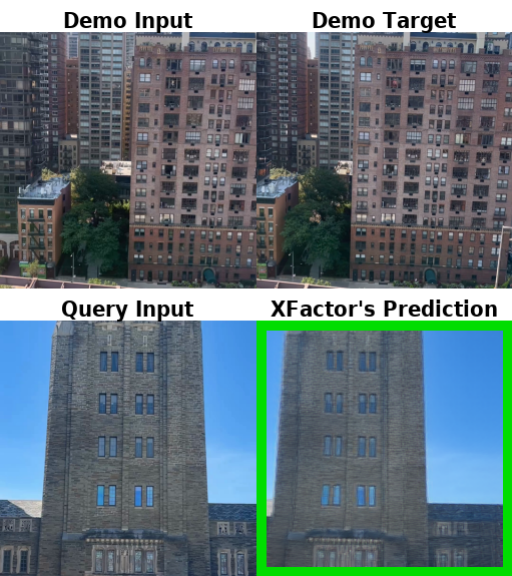}
        \caption{Setup: Demo: Standard $\rightarrow$ Query: Standard} 
        \label{fig:run1}
    \end{subfigure}
    \hspace{3mm}
    \begin{subfigure}[b]{0.35\textwidth}
        \centering
        \includegraphics[width=\textwidth]{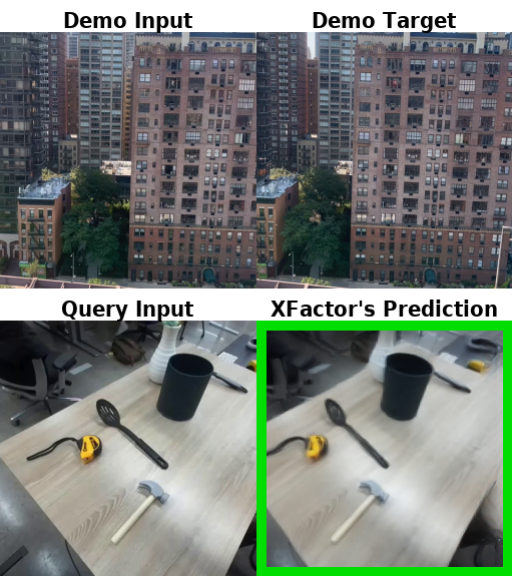}
        \caption{Setup: Demo: Standard $\rightarrow$ Query: Robotic} 
        \label{fig:run3}
    \end{subfigure}
    
    \vspace{1em} 
    
    \begin{subfigure}[b]{0.35\textwidth}
        \centering
        \includegraphics[width=\textwidth]{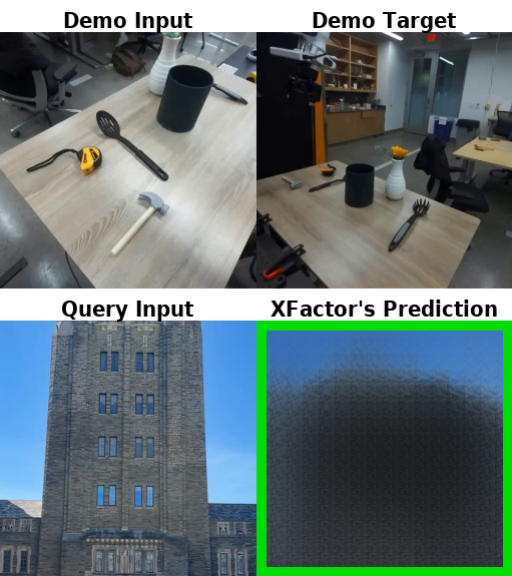}
        \caption{Setup: Demo: Robotic $\rightarrow$ Query: Standard} 
        \label{fig:run5}
    \end{subfigure}
    \hspace{3mm}
    \begin{subfigure}[b]{0.35\textwidth}
        \centering
        \includegraphics[width=\textwidth]{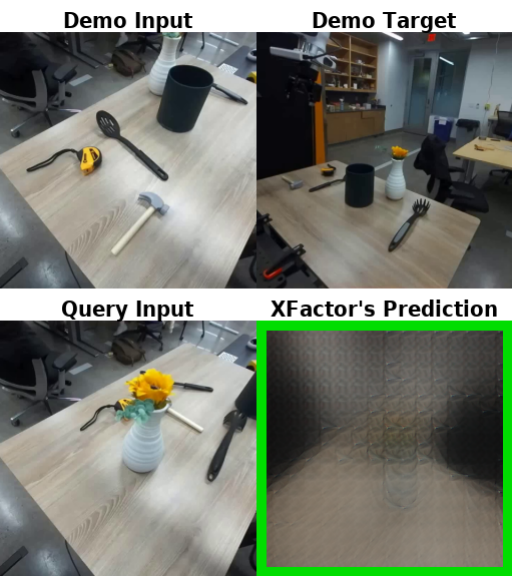}
        \caption{Setup: Demo: Robotic $\rightarrow$ Query: Robotic} 
        \label{fig:run7}
    \end{subfigure}
    
    \caption{\textbf{Qualitative failure mode analysis of XFactor.} We construct a 2x2 grid of cross-domain demo and query pairs. While XFactor easily extracts and applies standard, human-captured camera motions (a, b), it completely collapses when forced to process the extreme roll, pitch, and yaw typical of robotic side-camera setups (c, d). }
    \label{fig:xfactor_debug}
\end{figure}

The results, illustrated in Figure~\ref{fig:xfactor_debug}, are consistent with this hypothesis. When the demo transformation is sourced from the standard outdoor dataset (Figure~\ref{fig:xfactor_debug}a, b), XFactor successfully applies the motion to both outdoor and indoor query images. However, when the demo transformation is sourced from the indoor robotic cameras (Figure~\ref{fig:xfactor_debug}c, d), the failure appears to be driven by the camera motion rather than by the visual appearance of robotic scenes.

We attribute this failure to a gap between training and test rotation distributions. XFactor is trained on large scene- and object-level video datasets in which human camera operators rarely induce extreme roll or steep pitch. In contrast, the robotic demo used in Figures~\ref{fig:xfactor_debug}c and d involves a relative rotation of roughly 80° yaw, 75° pitch, and 30° roll, a combination that is likely underrepresented in XFactor's training data.

\newpage
\section{A Deep Dive to Video Analogy}

\begin{figure}[t]
\centering
\includegraphics[width=\textwidth]{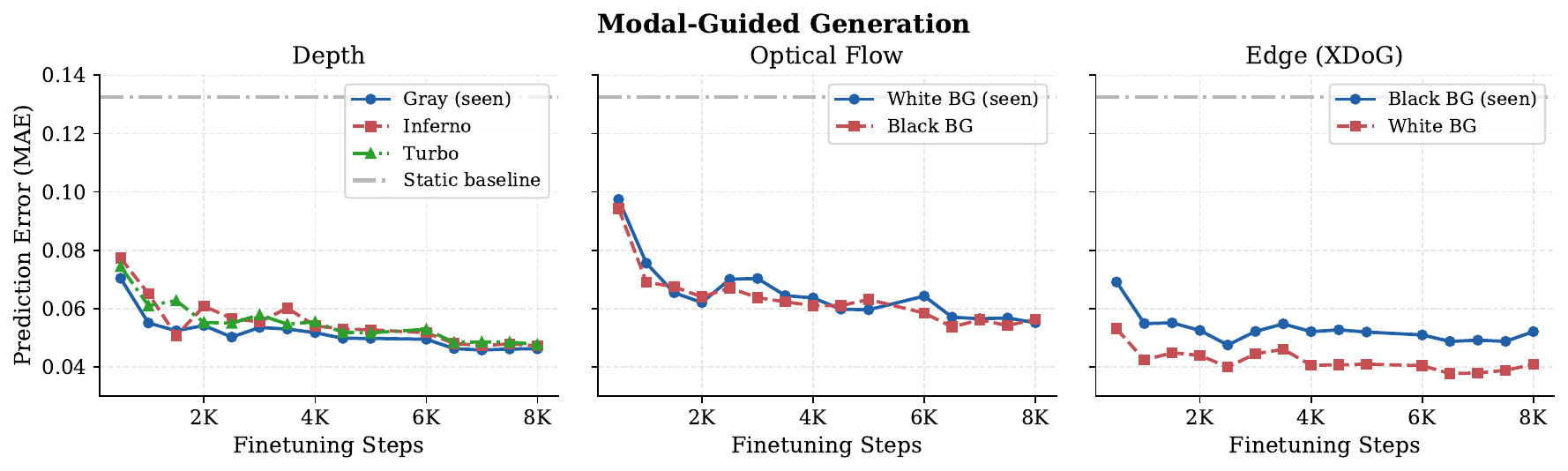}
\vspace{-6mm}
\caption{
\textbf{Robust perception across modality shifts: in-distribution vs. held-out guidance. }
Per-checkpoint MAE for modal-guided RGB generation under varying prompt styles. 
\model~successfully consumes out-of-distribution renderings as structural guidance. Left: Edge-guided generation works comparably well with seen (Black BG) and unseen (White BG) edge maps. Middle: Depth-guided generation maintains low error when prompted with unseen Inferno and Turbo colormaps, closely tracking the seen grayscale baseline. Right: Optical flow-guided generation adapts seamlessly to unseen black backgrounds. 
For reference, the `Static baseline' represents the MAE obtained by trivially repeating the first frame of the ground-truth video.
}
\label{fig:perceptual_invariance}
\end{figure}

\subsection{Additional result for task internalization}
\label{subsec:task_int_exp_flow_and_depth}

\begin{table}[thbp]
\centering
\caption{Quantitative effect of task internalization under the repeated first-frame hint ($F_0$ repeated), and its mitigation. Both directions are held out from training; the finetuning data contains neither depth nor optical
flow. \textbf{Best} results are bolded. $n{=}55$ sequences per cell.}
\label{tab:task_internalization}
\small
\setlength{\tabcolsep}{6pt}
\begin{tabular}{l cc cc}
\toprule
& \multicolumn{2}{c}{\textbf{Depth-to-Flow}} & \multicolumn{2}{c}{\textbf{Flow-to-Depth}} \\
\cmidrule(lr){2-3} \cmidrule(lr){4-5}
\textbf{Method}
& \textbf{MAE $\downarrow$} & \textbf{SSIM $\uparrow$}
& \textbf{MAE $\downarrow$} & \textbf{SSIM $\uparrow$} \\
\midrule
\model~\tiny{(base)}
& 0.375 & 0.335
& 0.333 & 0.257 \\
\cellcolor{lightblue}\model~\tiny{(+ finetuning)}
& \cellcolor{lightblue}\textbf{0.142} & \cellcolor{lightblue}\textbf{0.818}
& \cellcolor{lightblue}\textbf{0.108} & \cellcolor{lightblue}\textbf{0.793} \\
\bottomrule
\end{tabular}
\end{table}

Under the repeated first-frame hint, the base model produces an RGB copy rather than the demanded modality, giving an MAE between their outputs and targets of $0.375$ and SSIM of $0.335$ on depth-to-flow and $0.333$/$0.257$ on flow-to-depth. After finetuning, error (against their target) falls by $62\%$ and $68\%$ respectively, to $0.142$ and $0.108$, while SSIM rises to $0.818$ and $0.793$.

\subsection{Seeing through the style: \textit{Perceptual Invariance}}
\label{subsec:perceptual_invariance}

We ask: \textit{is the model's ability to comprehend input prompts brittle to visual styling shifts?}

\noindent \textbf{Setup.}
To evaluate this, we test the model on modal-guided video generation (predicting an RGB video from its first frame and a modal guidance video) but deliberately perturb the rendering style of the guidance. Specifically, we prompt the model with modalities in visual conventions it has never seen: depth maps rendered in Inferno and Turbo colormaps (instead of the training-standard grayscale), optical flow on a black background (instead of the training-standard Middlebury white), and edge maps on a white background (instead of the training-standard black). We then measure the prediction error (MAE) of the resulting generated RGB videos with the groundtruth video.

\noindent \textbf{Findings.}
The empirical results demonstrate perceptual robustness (Figure~\ref{fig:perceptual_invariance}). Despite the drastic pixel-level domain shifts in the guidance videos, the model consistently executes the correct structural generation. For depth-guided generation, the prediction error for unseen Inferno and Turbo prompts closely tracks the baseline grayscale performance. Similarly, optical flow guidance remains robust when inverted to a black background, and edge-guided generation succeeds regardless of the background polarity. We characterize this capability as perceptual invariance.

\noindent \textbf{Implications.}
The model's perceptual robustness provides deeper insight into how video analogies are processed within the canvas. We hypothesize that during the encoding phase, the model decouples high-frequency spatiotemporal structure (e.g., motion boundaries, geometric gradients) from low-frequency semantic styling (the absolute RGB colormap). It can essentially "see through" the color palette to extract the underlying geometric intent. For in-context learning frameworks, this implies that input guidance can afford to be highly diverse and out-of-distribution.

\newpage
\section{Task Taxonomy}
\label{app:task_taxonomy}

\subsection{Camera Pose}
\subsubsection{Training tasks.}
We organise camera-pose training into two task families.

\textbf{Camera motion transfer (1):} In \task{camera\_motion\_transfer}, the model is asked to generate a continuous video sequence along a target camera trajectory given (a) a single static frame to anchor appearance and (b) a demonstration of the same trajectory on a different scene. We curate this from two complementary sources. Layout is $2{\times}2$.
(i)CamCloneMaster~\citep{luo2025camclonemaster}: each demo and query row pairs a frozen first frame (left cell) with the corresponding full video (right cell); the demo and query rows are drawn from different scenes that share the same Unreal Engine 5-rendered camera trajectory, so the model must transfer the demonstrated motion to the query's appearance. 
(ii) SpatialVID~\citep{wang2025spatialvid}: we precompute a per-window descriptor of the camera path from the dataset's COLMAP poses and match windows that follow approximately the same trajectory across different real-world scenes. The matched pair is then assembled into the same layout. 

\textbf{Camera reangling (2):} The model is asked to synthesise the same scene as observed from a target camera placement $B$ given a video at an initial placement $A$ and a demonstration of the $A{\to}B$ correspondence. We use synchronised multi-camera real-world recordings from BridgeData-v2~\citep{walke2023bridgedata}, DROID~\citep{khazatsky2024droid}, and synthetic synchronised recordings from SynCamMaster~\citep{bai2024syncammaster}.
(i) \task{camera\_view\_transfer}: the demo row shows the same scene from two camera placements side-by-side as \texttt{[demo@A $\mid$ demo@B]}; the query row shows another scene from the same pair of placements as \texttt{[query@A $\mid$ query@B]}, where the right cell is the target the model must generate. Layout: $2{\times}2$.
(ii) \task{repose\_hinted}: we expand the pair task to a $2{\times}3$ grid by inserting a frozen first-frame at placement $B$ as a static ``hint'' between the two views, giving the layout \texttt{[A $\mid$ hint($B[0]$) $\mid$ B]} per row. The hint anchors appearance at the target placement so the model only has to recover the temporal motion, and provides a graded curriculum from the easier hinted variant to the harder unhinted \task{camera\_view\_transfer}.


\subsubsection{Evaluation tasks.}
For evaluation, we consider \task{camera\_view\_transfer} and \task{camera\_motion\_transfer}.
For \task{camera\_view\_transfer}, we measure generalisation to unseen environments, we hold out a small set of physical environments at the source level and exclude every scene, session, and trajectory belonging to them from training. From Bridge we hold out two environments, \texttt{toysink2\_bww} and \texttt{tabletop\_dark\_wood}; from DROID we hold out five buildings, \texttt{Bayes -- Kitchen 3}, \texttt{AHG 1st floor kitchen}, \texttt{Bayes -- ICMS}, \texttt{Bayes -- Kitchen Level 1}, and \texttt{Smith Hall 121}. The held-out evaluation set therefore measures whether the model can reangle a query scene captured in a physical environment it has never seen during training. 
For \task{camera\_motion\_transfer}, we use the 55 evaluation subset data and 100 drone subset data from DL3DV~\citep{ling2024dl3dv}.

\subsection{Multimodal}

%
%
%
%

\providecolor{rgbvidcol}{HTML}{C0392B}
\providecolor{modvidcol}{HTML}{808080}
\providecommand{\rgbvid}{{\color{rgbvidcol}\faVideo}}
\providecommand{\modvid}[1]{#1\,{\color{modvidcol}\faVideo}}
\providecommand{\rgbimg}[1]{#1\,{\color{rgbvidcol}\faImage}}
\providecommand{\modimg}[1]{#1\,{\color{modvidcol}\faImage}}

\begin{table*}[t]
\centering
\footnotesize
\label{tab:task_taxonomy_modal}
\caption{
\textbf{Per-task schema for Modal-Guided Generation.}
The demo and query rows share the same cell schema; the \emph{rightmost cell} is the model's target (present in the demo, absent and predicted in the query).
\textbf{Convention:} {\rgbvid} (red film icon) = \emph{RGB video}; {\modvid{$\cdot$}} (gray film icon, prefixed with the modality name) = \emph{modality video}. Analogously, {\rgbimg{}} = \emph{RGB image} (single frame) and {\modimg{$\cdot$}} = \emph{modality image}.
}
\label{tab:task_taxonomy_v2_modal}
\renewcommand{\arraystretch}{1.25}
\setlength{\tabcolsep}{5pt}
\begin{tabular}{l l p{7cm}}
\toprule
\textbf{Task} & \textbf{Row Schema} & \textbf{Notes} \\
\midrule
\multicolumn{3}{l}{\textit{\textbf{Training tasks}}} \\
\midrule
\task{est\_depth}        & [\rgbvid $\mid$ \modvid{depth}]                          &  \\
\task{est\_normal}       & [\rgbvid $\mid$ \modvid{normal}]                         &  \\
\task{est\_semantic}     & [\rgbvid $\mid$ \modvid{semantic}]                       & per-pixel semantic class (NYU-40 palette). \\
\task{est\_flow}         & [\rgbvid $\mid$ \modvid{flow}]                           &  \\
\task{est\_point}        & [\rgbvid $\mid$ \rgbimg{points}$^*$ $\mid$ \rgbvid$^\dagger$] & $^*$static dots on a black image frame, indicating the initial points to be tracked.\ \ $^\dagger$RGB video with point tracks overlaid. \\
\task{multimodal\_quad}  & [\rgbvid $\mid$ \modvid{quad}$^*$]                           & $^*$nested 2$\times$2 modality grid with four modalities each occupying a quadrant \\
\addlinespace[2pt]
\task{gen\_depth\_guided}    & [\rgbimg{} $\mid$ \modvid{depth} $\mid$ \rgbvid]       & --- \\
\task{gen\_normal\_guided}   & [\rgbimg{} $\mid$ \modvid{normal} $\mid$ \rgbvid]      & --- \\
\task{gen\_edge\_guided}     & [\rgbimg{} $\mid$ \modvid{edge} $\mid$ \rgbvid]        & --- \\
\task{gen\_position\_guided} & [\rgbimg{} $\mid$ \modvid{XYZ}$^*$ $\mid$ \rgbvid]    & $^*$a video where each pixel's color encodes the 3D location of the surface it shows. \\
\addlinespace[2pt]
\task{complete\_seg}   & [\rgbvid$^*$ $\mid$ \rgbvid]         & $^*$selected objects are masked (black) throughout the video except in \emph{one keyframe} per object where each is revealed; the model must reconstruct the full appearance and trajectory of every masked object from that single visible frame. \\
\task{track\_seg}          & [\rgbvid $\mid$ \modimg{hint}$^*$ $\mid$ \modvid{masklet}$^\dagger$] & $^*$an image where the target object is hinted with cross or scribbles.\ \ $^\dagger$an RGB video overlaid with the target object's contour or mask on every frame. \\
\midrule
\multicolumn{3}{l}{\textit{\textbf{Held-out evaluation tasks}}} \\
\midrule
\task{gen\_flow\_guided}   & [\rgbimg{} $\mid$ \modvid{flow viz} $\mid$ \rgbvid]          & --- \\
\task{gen\_point\_guided}  & [\rgbimg{} $\mid$ points\modvid{}$^*$ $\mid$ \rgbvid]             & $^*$~moving dots on a black frame. \\
\task{gen\_event\_guided}  & [\rgbimg{}$^*$ $\mid$ \modvid{event viz} $\mid$ \rgbvid]  & $^*$the first and last images are given for video interpolation.  \\
\bottomrule
\end{tabular}
\end{table*}

\begin{table*}[t]
\centering
\footnotesize
\label{tab:task_taxonomy_manipulation}
\caption{\textbf{Per-task schema for Video Manipulation.}
Demo and query rows share the same cell schema; the \emph{rightmost cell} is the model's target.
Demo and query use independently sampled degradation parameters, so the model must recover the operation from the demo row rather than memorize a fixed setting.
Held-out tasks test new operation families or unseen degradations.}
\label{tab:task_taxonomy_v2_manip}
\renewcommand{\arraystretch}{1.25}
\setlength{\tabcolsep}{5pt}
\begin{tabular}{l l p{4.5cm}}
\toprule
\textbf{Task} & \textbf{Row schema} & \textbf{Notes} \\
\midrule
\multicolumn{3}{l}{\textit{\textbf{Training tasks}}} \\
\midrule
\task{zoom-in\_static}        & [full $\mid$ crop]      & Static zoom-in to the same region \\
\task{zoom-in\_animated}      & [full $\mid$ dolly-in]  & Animated zoom-in to the same region \\
\task{zoom-out\_static}       & [crop $\mid$ full]      & --- \\
\task{zoom-out\_animated}     & [crop $\mid$ dolly-out] & --- \\
\addlinespace[2pt]
\task{change\_params}             & [orig $\mid$ degraded]  & Samples a tuning curve over time and applies it to the original image; variants include saturation, temperature, tone, and chromatic magnitude \\
\addlinespace[2pt]
\task{grid\_warp}             & [warped $\mid$ orig]    & shared Gaussian-bump field \\
\task{shuffle\_strip\_solve}  & [shuffled $\mid$ orig]  & 1D strips, per-row permutation \\
\task{shuffle\_strips\_apply} & [orig $\mid$ shuffled]  & 1D strips, shared permutation \\
\task{shuffle\_temporal}      & [shuffled $\mid$ orig] / [orig $\mid$ shuffled] & restore + apply variants \\
\midrule
\multicolumn{3}{l}{\textit{\textbf{Held-out evaluation tasks}}} \\
\midrule
\task{3d\_inpaint}            & [masked $\mid$ orig]      & 10--20 black space-time cubes as masks \\
\task{3d\_outpaint}           & [border-blk $\mid$ orig]  & 15--30 black space-time cubes as masks; the $(1{-}\text{mask})$ region is blacked out \\
\addlinespace[2pt]
\task{deblur\_spatial}        & [blurred $\mid$ sharp]    & {spatial blur} \\
\task{deblur\_motion}         & [blurred $\mid$ sharp]    & {motion-blur kernel} \\
\task{deblur\_temporal}       & [blurred $\mid$ sharp]    & {temporal averaging} \\
\addlinespace[2pt]
\task{temporal\_interp\_easy}  & [$N$ frames $\mid$ full]  & 16 random frames given for interpolation \\
\task{temporal\_interp\_hard}   & [$N$ frames $\mid$ full]  & 8 random frames given for interpolation \\
\addlinespace[2pt]
\task{jigsaw\_solve}          & [shuffled-2D $\mid$ orig] & {2D tile grid} (train was 1D strips) \\
\task{jigsaw\_apply}          & [orig $\mid$ shuffled-2D] & {2D tile grid}, shared perm \\
\addlinespace[2pt]
\task{low\_light}             & [dark+noise $\mid$ orig]  & $\gamma$-darken + Gaussian noise \\
\task{temporal\_flip}         & [reversed $\mid$ orig]    & {global temporal symmetry} \\
\task{horizontal\_flip}       & [mirrored $\mid$ orig]    & {global spatial symmetry} \\
\bottomrule
\end{tabular}
\end{table*}

Table~\ref{tab:task_taxonomy_v2_modal} summarizes the multimodal task taxonomy for both the 20 training tasks and 3 testing tasks.

\subsubsection{Training tasks.}
Training data is curated from three complementary sources: SpatialVID~\citep{wang2025spatialvid}, 
SA-V~\citep{kirillov2023segment}, 
and  
Hypersim~\citep{roberts2021hypersim}.
Per-pixel modalities for SpatialVID are produced by off-the-shelf estimators: Video Depth Anything~\citep{chen2025video} for depth, WAFT~\citep{wang2025waft} for optical flow, OneFormer~\citep{jain2023oneformer} for semantic segmentation, 
DSINE~\citep{bae2024rethinking} for surface normals, and AllTracker~\citep{harley2025alltracker} for point trajectories. 

\textbf{SpatialVID --- modality estimation (3):} The model is asked to recover a pixel-level modality from RGB. Layout: $2{\times}2$ \texttt{[RGB $\mid$ modality] $\times$ 2 rows}, with demo and query drawn from different scenes. We include \task{est\_depth}, \task{est\_normal}, and \task{est\_flow}.

\textbf{SpatialVID --- modality-guided RGB generation (3):} The model is asked to synthesise an RGB video given a frozen first frame (for appearance) and a per-frame modality video (for geometry/motion). Layout: $2{\times}3$ \texttt{[frozen first frame $\mid$ modality video $\mid$ RGB video] $\times$ 2 rows}. We include \task{gen\_depth\_guided}, \task{gen\_normal\_guided}, and \task{gen\_edge\_guided}, where the edge variant computes Sobel/XDoG/Canny edges on the fly with the method drawn per-sample from a fixed mixture.

\textbf{SpatialVID --- multimodal quad (1):} A $2{\times}2$ layout where the right cell is itself a nested $2{\times}2$ grid containing four modalities of the same scene (RGB, depth, normal, flow), each rendered as a $240{\times}240$ block. The task forces the model to consume and reproduce multiple modalities of the same scene jointly, and exposes the cross-modality consistency the demo and query rows must share. We refer to it as \task{multimodal\_quad}.

\textbf{SpatialVID --- point estimation (1):} In \task{est\_point}, the model is asked to track a sparse set of points across an RGB video. We sample $N \in \{5, 10, 20, 40\}$ trackable points per video using AllTracker, with semantic-aware filtering (visibility ${>}0.7$, confidence ${>}0.7$, in-frame fraction ${\ge}80\%$, sky/road/earth excluded via OneFormer's ADE20K labels, person/car preferred when available) so that the chosen points lie on physically meaningful, persistent surfaces rather than on the sky or distant background. The points are rendered as small colored dots (radius $8$--$12$, jittered colors). Layout: $2{\times}3$ \texttt{[RGB video $\mid$ static initial dots on black $\mid$ dots overlaid on RGB video] $\times$ 2 rows}. 

\textbf{Hypersim --- indoor multimodal (6):} Hypersim provides diverse, per-pixel modalities, including RGB, depth, surface normal in camera and world frames, semantic, world-space position, and instance) on synthetic indoor scenes. We sample 21 keyframes per (scene, camera) and repeat each keyframe four times to yield $82$ frames. We define four $2{\times}2$ tasks: \task{est\_depth}; \task{est\_normal} (with camera- and world-frame variants); 
\task{est\_semantic} (NYU-40 palette); 
and \task{complete\_seg}, in which we pick $4$--$5$ instance-segmented objects that are visible in at least $18$ of the $21$ keyframes and render each object unmasked in only one keyframe and mask the objects in all other frames to black, requiring the model to reconstruct the full video from sparse single-frame appearance observations. We additionally define two $2{\times}3$-layout tasks: \task{gen\_position\_guided} (frozen first-frame world-space XYZ rendered as RGB serves as guidance for the full position video; 
and \task{gen\_depth\_guided}.


\textbf{SA-V --- video segmentation (1):} SA-V provides ${\sim}190$K masklet annotations over real-world video. We use a $2{\times}3$ layout for
\task{track\_seg} (left: RGB video; mid: black video whose first frame carries a hint of the masklet, one of \{coloured semi-transparent mask, contour outline, three cross markers, bounding box, scribble\} drawn per-sample; right: the masklet rendered as a coloured overlay on every frame, the model's target).

\subsubsection{Evaluation tasks.}
We evaluate the model on two held-out modality-guided RGB-generation tasks built from the SpatialVID source pool, using the same $2{\times}3$ layout as the corresponding training tasks but with modalities the model has \emph{never} been trained to consume as guidance:
(i) \task{gen\_flow\_guided}: the middle column is a per-frame optical-flow visualisation produced by WAFT, and the model must synthesise the corresponding RGB video;
(ii) \task{gen\_point\_guided}: the middle column shows $N=16$ moving points (from AllTracker, with the same semantic-aware filtering used at training time) on black, and the model must produce the matching RGB video where the points implicitly anchor a sparse, deterministic motion field. 
(iii) \task{gen\_event\_guided}: the left video is sparse RGB frame videos; the middle column is a event camera data visualization, and the model must synthesise the corresponding RGB video.

For \task{gen\_flow\_guided}, we compare with Go-With-The-Flow~\citep{burgert2025go} and FloVD~\citep{jin2025flovd}. FloVD is a generative model based on CogVideoX that contains a flow synthesis module and a flow-conditioned video synthesis module (FVSM). We use its FVSM module for evaluation. 

For \task{gen\_event\_guided}, we compare with CBMNet~\citep{kim2023event} and RE-VDM~\citep{chen2025repurposing}, both taking first and end frame video and raw event camera data (CBMNet gets binned grids) for video frame interpolation. 
For our model, it takes in two boundary anchor frames, identical to the frames the event baselines are given, with all intermediate frames blanked. Specifically, we prepare the input video by "placing frame 0 at idx 0, and frame 12 at idx 12; while other 79 frames are black".

\subsection{Video Manipulation}

Table~\ref{tab:task_taxonomy_v2_manip} summarize the video manipulation taxonomy for both the training and testing tasks.

\subsubsection{Training tasks.}
We consider three categories of tasks and enumerate in total $9$ tasks.

\textbf{Zoom tasks (4):} We consider four variants spanning the cross product of \{zoom-in, zoom-out\} $\times$ \{static, animated\}: \task{zoom-in\_static}, \task{zoom-in\_animated}, \task{zoom-out\_static}, \task{zoom-out\_animated}, all sourced from the 4K ($3840{\times}2160$) data from DL3DV~\citep{ling2024dl3dv}. 
For every scene we deterministically sample two nested square regions in the source frame: an outer ``wide'' square of side $S \in [720,\, 1920]$\,px (i.e.\ up to $4{\times}$ the close-up size, capped by the frame's shorter dimension) and an inner $480{\times}480$ ``close-up'' bounding box placed at a random offset inside the wide square. Both the outer square and the inner bounding box are \emph{shared} between the demo and query rows, so the model must localise the close-up region by comparing the two rows. The two views are rendered as $480{\times}480$ blocks: the wide view is the outer square downsampled to $480{\times}480$, and the close-up view is the inner bounding box at native pixel scale. In the \emph{static} variants both rows display these two views side by side as still cells, ordered \texttt{[full $\mid$ crop]} for zoom-in and \texttt{[crop $\mid$ full]} for zoom-out, with no temporal zoom motion. In the \emph{animated} variants one of the two cells is replaced by a synthetic dolly-zoom: at every frame we recompute the crop window size, smoothly sweeping it from the outer square to the inner bounding box (zoom-in) or the reverse (zoom-out), and resize back to $480{\times}480$.

\textbf{Low-level image tasks (1):} We consider a \task{change\_params} task in which low-level image attributes are altered with a smooth time-varying parameter that is shared between the demo and query rows, so the model must recover the underlying control signal from pixel differences alone. The control trajectory $p(t) \in [0,1]$, $t = 1, \dots, 82$, is generated as follows: we draw $2$--$3$ random anchor values uniformly from $\{\tfrac{1}{20}, \tfrac{2}{20}, \dots, \tfrac{19}{20}\}$, prepend the two endpoints $\{0, 1\}$ to guarantee that both extremes are visited, randomly shuffle the combined list to determine the visit order, then concatenate equal-length segments interpolating between consecutive anchors using one of four randomly chosen easing modes (linear, ease-in, ease-out, ease-in-out). Each task maps $p(t)$ to a different image-domain parameter: (i) saturation, a vectorised RGB saturation factor in $[0, 2]$ blending each frame with its luminance map; (ii) temperature, a warm/cool shift implemented by per-frame R/B channel offsets up to $\pm 40$; (iii) tone, a per-frame gamma correction with $\gamma \in [0.4,\, 2.8]$ applied via cached LUTs; (iv) focus, a Gaussian blur whose kernel size sweeps from $1$ to $51$ pixels along the trajectory; and (v) chromatic aberration, per-frame integer translations of the R and B channels by up to $\pm 20$\,px along a smooth 2D path, simulating chromatic aberration as a controllable channel-shift effect. For all five, the same trajectory is used for both rows, so the demo row implicitly defines the time-varying parameter that the query row must match.

\textbf{Spatiotemporal correspondence tasks (4):} These tasks geometrically or temporally permute the input and require the model to either invert the permutation (\emph{restore}) or replicate it (\emph{apply}). We consider:
(i) \task{grid\_warp}, which composes $2$--$6$ Gaussian ``bumps'' on a
$720{\times}720$ canvas with sliding, rotational, or pulsing behaviours and produces a smooth shared spatial deformation field;
(ii)
\task{shuffle\_strip\_solve}, which slices each $480{\times}480$ frame into $3$--$6$ horizontal \emph{or} vertical strips and applies an independent non-identity permutation per row in a \texttt{[shuffled $\mid$ original]} layout, so the model must
learn to unshuffle by attending to the demo row; (iii)
\task{shuffle\_strips\_apply}, which uses the \emph{same} strip permutation for
both rows in a \texttt{[original $\mid$ shuffled]} layout, so the model must
learn to apply the permutation demonstrated in the demo row; and (iv)
\task{shuffle\_temporal}, which permutes frames along the time axis with two
hash-selected sub-variants: \emph{restore} (independent shuffle per row) and
\emph{apply} (shared shuffle across rows). We note that
\task{shuffle\_strip\_solve}/\task{shuffle\_strips\_apply} differ from the classical Jigsaw
pretext task in that our strips are one-dimensional bands (horizontal or
vertical only), whereas Jigsaw permutes a 2D tile grid, and our setup
explicitly provides a paired demo--query row rather than a single-image
classification target.

\subsubsection{Evaluation tasks.}
We use the $55$ scenes from the DL3DV evaluation split and apply $15$ video-processing tasks grouped into five categories. All evaluation tasks share the same $2{\times}2$ grid layout \texttt{[degraded $\mid$ original] $\times$ 2 rows}: each row pairs a degraded input (left) with the clean target (right), and the demo and query rows use independently sampled degradation parameters so that the model recovers the operation from the demo row rather than memorising a fixed setting.

\textbf{Inpainting (2):} Tasks that mask out parts of the video and require the model to fill them in.
(i) \task{3d\_inpaint}, $10$--$20$ black space-time cubes of size $50$--$150$\,px and $10$--$40$ frames placed at random $(x, y, t)$ offsets, producing time-varying occlusions;
(ii) \task{3d\_outpaint}, the inverse of \task{3d\_inpaint}: only $15$--$30$ space-time cubes show the original RGB content while everything else is black, so the model must reconstruct the full video from sparse 3D ``windows''. 

\textbf{Deblurring (3):} Tasks that degrade the input with a blur and require the model to recover sharp content.
(i) \task{deblur\_spatial}, a per-scene-randomised spatial blur drawn uniformly from \{Gaussian, box, median\} with kernel size $11$--$27$\,px;
(ii) \task{deblur\_motion}, a directional motion blur using a linear streak kernel of size $15$--$33$\,px at a uniformly sampled angle in $[0, 180^\circ)$;
(iii) \task{deblur\_temporal}, a temporal blur produced by averaging each frame with its $\pm 4$-frame neighbours (window size $9$, edges truncated), simulating long-exposure or low-frame-rate capture.

\textbf{Temporal interpolation (2):} Tasks that delete most frames and require the model to interpolate the missing ones. We sparsify the input by keeping exactly $N$ randomly chosen frames intact and zeroing out the rest, with two difficulty levels controlled by $N$:
(i) \task{temporal\_interp\_easy} ($N{=}16$);
(ii) \task{temporal\_interp\_hard} ($N{=}8$);
The set of visible frame indices is sampled independently per row.

\textbf{Spatial jigsaw (2):} Tasks that permute a $2{\times}2$ or $3{\times}3$ tile grid of each frame.
(i) \task{jigsaw\_solve}, where each row is shuffled with an independently sampled non-identity permutation in a \texttt{[shuffled $\mid$ original]} layout, so the model must invert the demonstrated permutation;
(ii) \task{jigsaw\_apply}, where both rows share the \emph{same} permutation in an \texttt{[original $\mid$ shuffled]} layout, so the model must replicate the demonstrated permutation. Unlike the strip-shuffle training tasks (which act on $1$D bands), these jigsaw tasks act on a true $2$D tile grid and serve as the canonical out-of-distribution counterpart.

\textbf{Photometric \& symmetry (3):} Three additional tasks covering low-level photometric correction and global symmetry operations.
(i) \task{low\_light}, a low-light enhancement task that gamma-darkens each frame with $\gamma \in [0.15,\, 0.30]$ (effective exponent $1/\gamma \in [3.3,\, 6.7]$) and adds Gaussian noise with $\sigma \in [10,\, 30]$;
(ii) \task{temporal\_flip}, which reverses the playback order of the input video, requiring the model to play the scene backwards;
(iii) \task{horizontal\_flip}, which mirrors each frame left-to-right, requiring the model to produce the left-right reflected video.

\subsection{Camera Pose Metrics}
\label{subsec:camera_pose_metrics}

We follow the formulation in CamI2V~\citep{zheng2024cami2v}.
Given a generated video of $S$ frames, we extract camera extrinsics using VGGT~\citep{wang2025vggt} and compare against the ground-truth camera trajectory. Both predicted and ground-truth trajectories are first converted to camera-to-world (c2w) poses, made relative to the first frame (setting frame~0 as identity), and independently scale-normalized by dividing all translations by the distance to the furthest camera from the origin:
\begin{equation}
    \hat{\mathbf{t}}_i = \frac{\mathbf{t}_i}{\max_{j} \|\mathbf{t}_j\|_2}, \quad i = 1, \ldots, S-1,
\end{equation}
where $\mathbf{t}_i$ denotes the translation of the $i$-th frame relative to frame~0. This normalization is applied independently to both GT and predicted trajectories, making the metrics scale-invariant.

\noindent \textbf{Rotation Error (RotErr).} The sum of geodesic (angular) distances between predicted and ground-truth rotations:
\begin{equation}
    \text{RotErr} = \sum_{i=0}^{S-1} \arccos\!\left( \frac{\text{tr}(\mathbf{R}_i^{\text{est}} {\mathbf{R}_i^{\text{gt}}}^\top) - 1}{2} \right),
\end{equation}
where $\mathbf{R}_i^{\text{est}}, \mathbf{R}_i^{\text{gt}} \in SO(3)$ are the predicted and ground-truth rotation matrices at frame $i$, and the argument of $\arccos$ is clipped to $[-1, 1]$.

\noindent \textbf{Translation Error (TransErr).} The sum of $\ell_2$ distances between predicted and ground-truth translations after scale normalization:
\begin{equation}
    \text{TransErr} = \sum_{i=0}^{S-1} \left\| \hat{\mathbf{t}}_i^{\text{est}} - \hat{\mathbf{t}}_i^{\text{gt}} \right\|_2.
\end{equation}

\noindent \textbf{Camera Motion Error (CamErr).} The sum of Frobenius norms of the $3 \times 4$ pose matrix differences:
\begin{equation}
    \text{CamErr} = \sum_{i=0}^{S-1} \left\| [\mathbf{R}_i^{\text{est}} \mid \hat{\mathbf{t}}_i^{\text{est}}] - [\mathbf{R}_i^{\text{gt}} \mid \hat{\mathbf{t}}_i^{\text{gt}}] \right\|_F.
\end{equation}

\noindent \task{Camera\_view\_transfer}.
For the camera view task, we treat the input image (pose~A) and the generated 
output image (target pose~B) as a two-frame sequence ($S{=}2$). Ground-truth 
extrinsics are obtained by running VGGT on the two real images from the demo pair. Predicted extrinsics are obtained by running VGGT on the query input image and the first frame of the generated output. We then apply the same RotErr, TransErr, and CamErr formulation with $S{=}2$. Since frame~0 is canonicalized to identity, only the relative pose of the generated view contributes to the error.

\noindent \task{Camera\_motion\_transfer}.
For the camera motion task, the ground-truth is the full $S{=}81$ frame trajectory 
from the reference video. Predicted extrinsics are extracted from the generated 
video using VGGT. Since all metrics are sums over $S$ frames, values are not 
directly comparable between the two tasks.

\section{Limitations}
\label{sec:limitations}

As the first work to explore video analogies as a unified framework for diverse tasks, we acknowledge several current limitations. We thoroughly detail these in Section~\ref{subsec:experients_video_manipulation}, which discusses test tasks our model struggles to perform (such as flipping and deblurring) , and Section~\ref{sec:deep_dive}, which characterizes bounds on generalization like task internalization.

Beyond these specific failure modes, our framework opens several promising avenues for future work. First, \textit{incorporating text conditioning}. In its current form, our model relies entirely on visual prompts; future work could investigate how integrating text-based task and video prompts might enhance the performance and generative quality of video analogies.  

Second, \textit{expanding task diversity}. Despite our best efforts, the diversity of our training tasks remains bounded. As noted in Section~\ref{sec:experiments} and Section~\ref{sec:deep_dive}, increasing task variety and training the model to generate a wider array of visual appearances are integral steps toward making the system more robust and generalizable.  

Finally, \textit{exploring diverse visual modalities}. Future research could extend this approach to other complex data structures, such as spectrograms, sonograms, 3D point clouds, sinograms, or volumetric 3D images (e.g., CT and MRI scans).





\end{document}